\documentclass[letterpaper]{article} 
\usepackage{aaai2027}  
\usepackage[hyphens]{url}  
\usepackage{graphicx} 
\usepackage{natbib}  
\usepackage{caption} 
\usepackage{algorithm}
\usepackage{algorithmic}
\usepackage{algorithm}
\usepackage{algorithmic}
\usepackage{subcaption}

\usepackage{tabularx}
\usepackage{newfloat}
\usepackage{listings}
\DeclareCaptionStyle{ruled}{labelfont=normalfont,labelsep=colon,strut=off} 
\floatstyle{ruled}
\newfloat{listing}{tb}{lst}{}
\floatname{listing}{Listing}

\usepackage{booktabs}

\newcommand{\src}{\text{src}}
\newcommand{\tar}{\text{tar}}

\usepackage{booktabs}

\usepackage[dvipsnames]{xcolor} 
\usepackage{pifont}             

\newcommand{\cmark}{\textcolor{Green}{\ding{51}}} 
\newcommand{\xmark}{\textcolor{Red}{\ding{55}}}    

\usepackage{pifont}
\usepackage{multirow}
\usepackage{amsmath}
\usepackage{amssymb}

\usepackage[table]{xcolor}  
\definecolor{RankFirst}{RGB}{255,242,204}  
\definecolor{RankSecond}{RGB}{255,204,153} 
\definecolor{RankThird}{RGB}{221,235,247} 
\definecolor{coralPink}{HTML}{ED028C}

\title{WhereEdit: Mask-aware Local Latent Editing for One-Step Image Editing}
\author{
    Ming Hu$^{1,2}$, Mingyu Dou$^{1,2}$, Jianfu Yin$^{1,2}$, Miaomiao Zhang$^{1,2}$, Cong Hu$^{4}$\\
    Yao Wang$^{3}$, Bingliang Hu$^{1}$, Quan Wang$^{1}$\\
}
\affiliations{
    $^1$ Xi'an Institute of Optics and Precision Mechanics, Chinese Academy of Sciences \\
    $^2$ University of Chinese Academy of Sciences \quad 
    $^3$ Xi'an Jiaotong University\\
   $^4$  Zhongnan Hospital of Wuhan University
}

\begin{document}

\maketitle

\begin{abstract}
Recent one-step text-to-image (T2I) models enable efficient image synthesis and provide new opportunities for real-time image editing. However, existing one-step editing methods primarily rely on text conditioning for semantic transformation, lacking explicit spatial control over \textit{where} to edit. More importantly, even when spatial constraints are introduced, these methods often struggle to achieve strong and stable semantic modifications within the target regions.
In this work, we revisit one-step image editing from a spatially controlled perspective and identify two key challenges: discovering editable regions and achieving effective localized semantic transformation. We reveal that existing methods perform global semantic transport, which limits high-intensity local editing under the one-step setting. To address this issue, we propose \textbf{WhereEdit}, a framework that reformulates one-step editing as localized adaptive editing. WhereEdit automatically identifies semantically relevant regions from internal model features and applies adaptive local modulation to enhance target-region editing while preserving non-target areas and structural consistency.
Experiments on the PIE-Bench benchmark demonstrate that WhereEdit consistently outperforms existing one-step image editing methods, achieving superior editing quality while maintaining the efficiency of one-step generation. Additional experiments with region-level supervision further highlight the importance of explicit spatial reasoning for high-quality one-step image editing.
\end{abstract}

\begin{links}
    \link{Code}{https://github.com/Xi-Mu-Yu/WhereEdit}
\end{links}

\section{Introduction}

Recent advances in one-step text-to-image (T2I) generation models have introduced a new paradigm for real-time visual synthesis~\cite{lu2026chordedit,nguyen2025swiftedit, liu2023instaflow, Dao2025SwiftBrushV2,Sauer2024ADD}. By distilling large-scale diffusion models~\cite{song2023consistency,lin2024sdxl}into compact one-step inference frameworks, these approaches dramatically reduce generation latency and make interactive image creation increasingly feasible in practical applications. However, this efficiency-oriented paradigm also raises a fundamental question: how can one-step T2I models be extended beyond image synthesis toward fine-grained, structured, and controllable image editing?

\begin{figure}[h]
\centering
\includegraphics[width=\columnwidth]{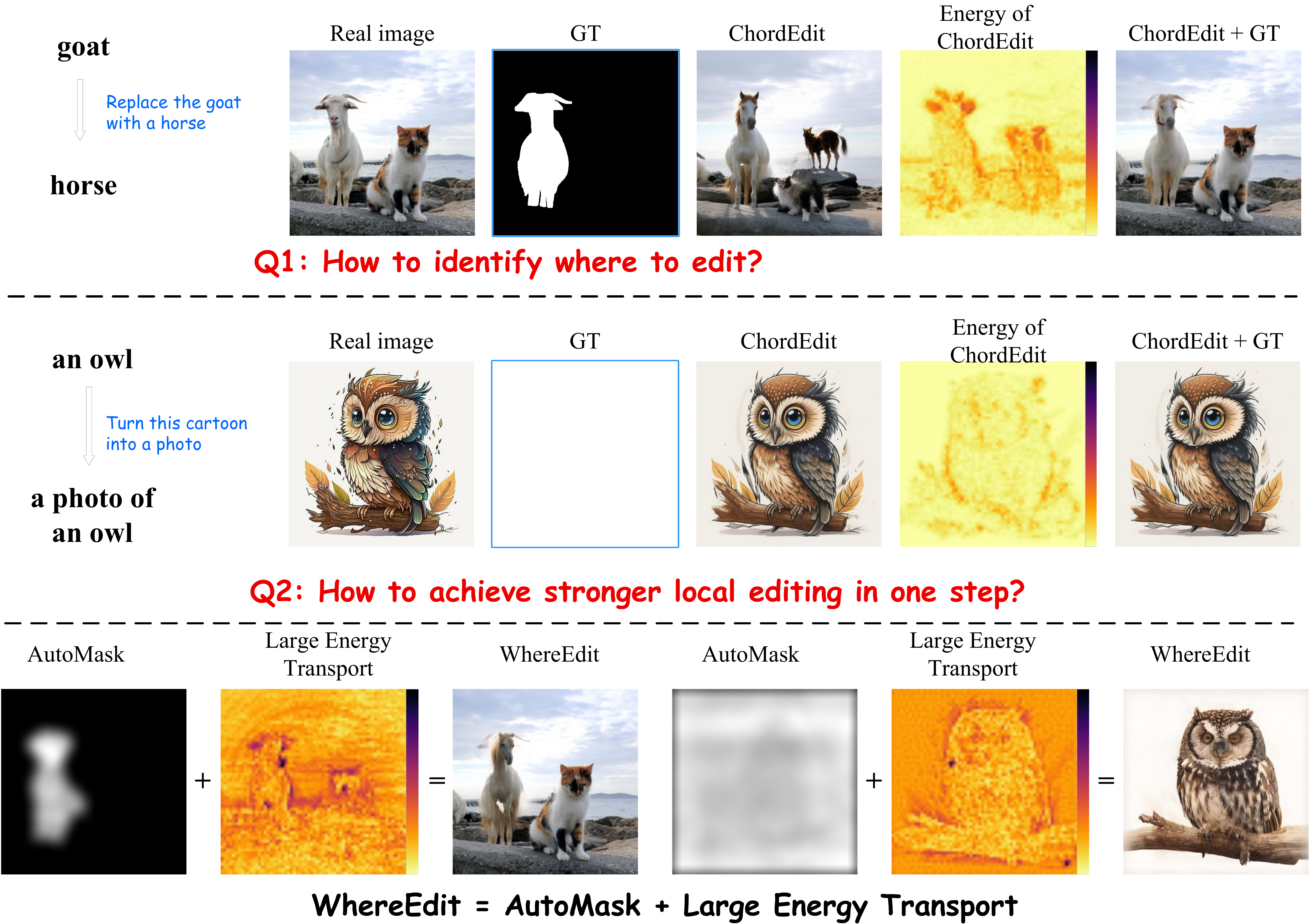}
\caption{The first row demonstrates that ChordEdit benefits significantly from GT mask guidance, highlighting the importance of explicit spatial constraints. The second row reveals its limitation caused by the intrinsic low-energy property, which struggles with large semantic changes. The last row presents our WhereEdit, which introduces AutoMask and large-energy transport to enable more effective editing, illustrating the motivations behind Q1 and Q2.}
\label{mitivation}
\end{figure}

Although recent studies have explored adapting one-step generative models for image editing~\cite{lu2026chordedit,nguyen2025swiftedit}, most existing approaches remain constrained by unconditional or text-conditioned generation paradigms, lacking explicit control over \textbf{where} modifications should occur. As illustrated in Fig.~\ref{mitivation}, we investigate this limitation in ChordEdit by introducing ground-truth (GT) masks as explicit spatial constraints. The substantial improvement in PSNR demonstrates that accurate spatial guidance can significantly enhance editing quality, revealing an important observation: existing one-step editors possess the ability to perform semantic transformation, but lack an effective mechanism to automatically identify the regions requiring modification. 
This motivates our first research question: \textbf{how can the editable regions be automatically discovered from the input image and editing instruction?}

Beyond localization, we further observe that even with GT masks providing ideal spatial supervision, existing approaches still struggle with large-scale semantic rewriting inside the target region. Specifically, the edited content often exhibits insufficient transformation strength and unstable local details, indicating that accurate region identification alone is insufficient for high-quality one-step editing. This leads to our second research question: \textbf{how can we achieve strong and stable semantic modification within a specified editing region?}

Based on these observations, we revisit one-step image editing from a spatially controlled generation perspective and formulate it as a mask-guided editing problem~\cite{hu2024specslice, hu2026fb, hu2025beta}. In this formulation, the text prompt provides semantic direction, while the spatial mask explicitly determines the scope of modification. The key challenge therefore shifts from merely learning \textbf{what} to edit, toward understanding \textbf{where} and \textbf{how strongly} the editing process should be applied.
To better understand this challenge, we analyze the underlying mechanism of existing one-step editing methods. Taking ChordEdit as an example, its editing process can be interpreted as estimating a global semantic displacement field between source and target conditions. Such a global transport mechanism enables smooth semantic transition across the entire latent space, but also introduces an inherent limitation for localized high-intensity editing. When spatial constraints are applied, the global displacement is simply truncated by the mask, resulting in reduced effective editing capacity within the target region. Consequently, large semantic transformations become difficult to achieve, especially under the one-step setting where iterative refinement is unavailable.

Motivated by this observation, we propose \textbf{WhereEdit}, a new framework that reformulates one-step image editing from global semantic transport into localized adaptive editing. Instead of passively restricting a global editing field, WhereEdit first discovers the semantically relevant editing region from internal model features and then performs concentrated editing within the identified area. Through adaptive local modulation, our approach enables stronger semantic rewriting while suppressing undesired modifications in non-target regions and maintaining structural consistency.
Extensive experiments on the PIE-Bench benchmark demonstrate that WhereEdit achieves SOTA performance among one-step image editing methods while preserving the efficiency advantage of one-step generation. Furthermore, when accurate region-level supervision is provided, our method obtains additional improvements, validating the importance of explicit spatial reasoning for high-quality one-step image editing.

\section{Related Work}

\subsection{Text-guided Editing with Diffusion/Flow Models}

With the rapid development of diffusion and flow-based generative models~\cite{song2020denoising, ho2020denoising, peng2025facm, geng2026mean}, text-guided image editing has become a dominant paradigm for controllable image manipulation. 
Early approaches typically rely on image inversion, mapping inputs into the latent space of pretrained generative models and performing edits through attention control, guidance, or latent optimization. 
Methods such as PnP~\cite{hertz2022prompt, tumanyan2023plug}, NPI~\cite{mokady2023null, npi25}, and ProxEdit~\cite{ju2023direct, proxedit24, edict23, aidi23, spdinv24} achieve high-quality editing but require iterative inversion and multi-step sampling, resulting in substantial computational costs. 
Recent fast generative models, including SD-Turbo~\cite{Sauer2024ADD}, InstaFlow~\cite{liu2023instaflow}, and SwiftBrush-v2~\cite{Dao2025SwiftBrushV2}, enable one- or few-step generation, but their accelerated generation dynamics introduce new challenges for precise and controllable editing.

\subsection{Training-free and Inversion-free Editing}

To improve efficiency, recent studies explore training-free and inversion-free editing by avoiding per-image inversion and additional training. 
Flow-based methods such as FlowEdit~\cite{kulikov2025flowedit} and InfEdit~\cite{xu2023inversion} construct editing directions through transport-based formulations, but still rely on multi-step sampling or trajectory estimation for stability. 
In the one-step setting~\cite{icd24, nguyen2025swiftedit, kawar2023imagic}, compressing the editing trajectory into a single update often leads to unstable optimization and degraded fidelity. 
ChordEdit~\cite{lu2026chordedit} provides an efficient semantic residual direction, but lacks an explicit target-attraction mechanism, limiting editing strength in challenging scenarios.

\subsection{Mask-guided Controlled Image Editing}

Spatial controllability is crucial for precise image editing. 
Mask-guided approaches introduce spatial constraints, such as user masks or automatically discovered regions, to localize modifications and preserve unedited content. 
Recent methods explore attention-based localization and region-aware optimization. 
LIME~\cite{Simsar_2025_WACV} discovers editable regions using cross-attention and intermediate features with attention regularization, while LocInv~\cite{tang2024locinv} reduces attention leakage through localization priors and attention alignment. 
MAG-Edit~\cite{mao2024mag} improves regional alignment by optimizing attention dominance within masks, and LayerEdit~\cite{fu2026layeredit} performs object-level decomposition for multi-object disentangled editing.

Despite their effectiveness, existing approaches generally rely on inversion, iterative sampling, external segmentation, or additional region discovery and optimization procedures, which introduce computational overhead and limit their applicability to one-step editing. 
In this work, WhereEdit addresses spatially controlled one-step editing by introducing a training-free and inversion-free attention-guided AutoMask mechanism. 
By aggregating attention responses from inserted target tokens and removed source tokens, AutoMask provides efficient spatial localization without additional clustering or optimization, while Amplified Conditional Transport enables stronger semantic manipulation within the one-step generation regime.

\begin{figure*}[h]
\centering
\includegraphics[width=\textwidth]{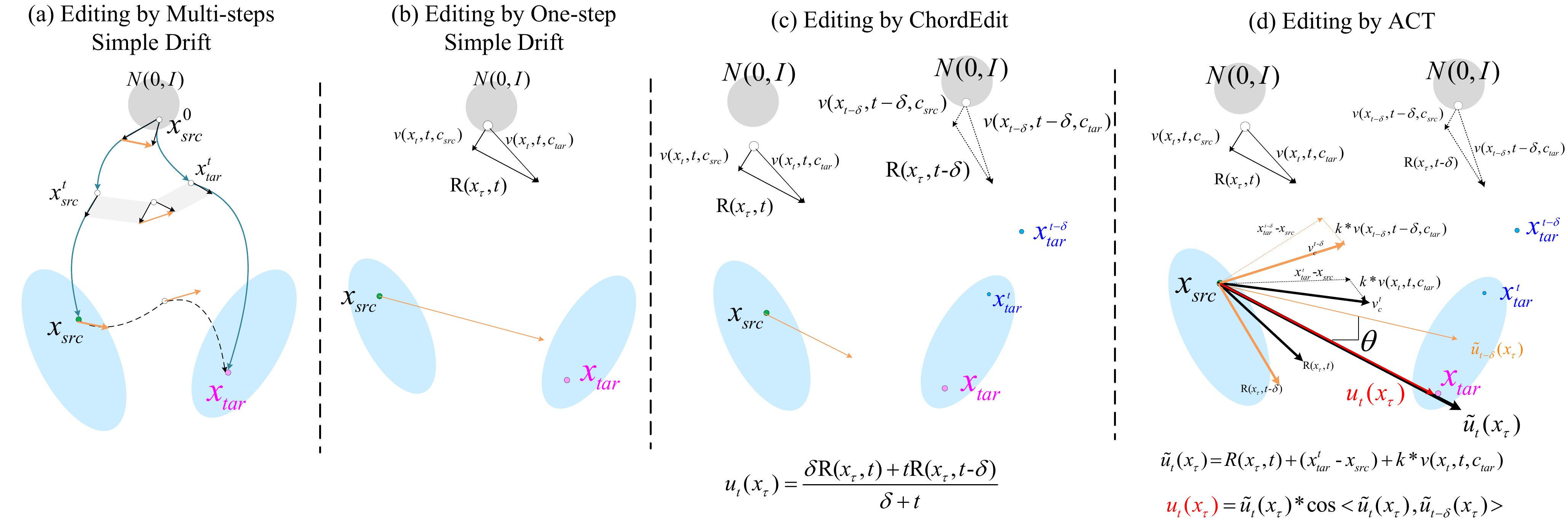}

\caption{\textbf{Comparison of editing field properties.}
(a) Multi-step diffusion: iterative updates produce a stable trajectory.
(b) Naive one-step transport: the raw conditional field lacks sufficient
target attraction and correction, leading to inaccurate target reaching.
(c) ChordEdit: temporal averaging produces a smooth and low-energy field,
achieving stable but conservative transport with limited large-edit capability.
(d) ACT (Ours): an amplified target-attraction component and consistency gate
provide stronger and corrected transport, enabling robust one-step editing
for large semantic changes.}
\label{fig:compare_editing}
\end{figure*}

\section{Method}

\label{sec:method}

We propose \textbf{WhereEdit}, a AutoMask-Guided one-step image editing framework.
WhereEdit addresses two coupled challenges in one-step editing:
(i)~\emph{where} to edit, by automatically localizing the target region from the input image and text instruction; and
(ii)~\emph{how strongly} to edit, by replacing the global low-magnitude energy transfer of prior transport-based editors with a localized large-magnitude energy shift inside the inferred region. 


\subsection{Amplified Conditional Transport}

We introduce Amplified Conditional Transport (ACT), a one-step editing
field that enhances conditional target attraction beyond naive conditional
transport. Instead of directly applying the difference between source and
target conditional predictions, ACT introduces an explicit attraction toward
the target conditional prediction, resulting in a state-dependent transport
field.

Let $x$ denote the current editing anchor and let
$\hat{x}_{0}^{\theta}$ denote the denoising prediction under condition $c$.
Using a shared noise bank
$\{\epsilon^{(i)}\}_{i=1}^{n}$, we define the conditional mean of denoising
predictions as
\begin{equation}
\hat{\mu}_{c}(t,x)
=
\frac{1}{n}
\sum_{i=1}^{n}
\hat{x}_{0}^{\theta}
(\alpha_t x+\sigma_t\epsilon^{(i)},t,c).
\end{equation}

The source and target conditional means are denoted as
$\hat{\mu}_{\mathrm{src}}$ and $\hat{\mu}_{\mathrm{tar}}$, respectively.
Since the same noise samples are shared across different conditions, the
resulting estimator forms a paired Monte Carlo estimate, which reduces the
variance when measuring the conditional transition between source and target
predictions.
A straightforward conditional transport direction can be obtained by
directly subtracting the source and target conditional predictions:
\begin{equation}
\mathcal{R}_{\mathrm{naive}}
=
\hat{\mu}_{\mathrm{tar}}
-
\hat{\mu}_{\mathrm{src}} .
\end{equation}

Although this formulation captures the average semantic difference between
the two conditions, it does not explicitly consider the current editing
state $x$ or provide an additional attraction mechanism toward the target
conditional prediction.
To enhance target-oriented editing, ACT introduces an amplification
coefficient $k\geq0$ and defines the one-step editing field as
\begin{equation}
\hat{u}^{\mathrm{ACT}}_{k}
=
(1+k)\hat{\mu}_{\mathrm{tar}}
-
\hat{\mu}_{\mathrm{src}}
-x .
\end{equation}

Equivalently, ACT can be interpreted as a displacement toward a dynamically
constructed conditional attractor:
\begin{equation}
x_k^{\star}(t,x)
=
(1+k)\hat{\mu}_{\mathrm{tar}}(t,x)
-
\hat{\mu}_{\mathrm{src}}(t,x),
\end{equation}

such that
\begin{equation}
\hat{u}^{\mathrm{ACT}}_{k}
=
x_k^{\star}(t,x)-x .
\end{equation}

Unlike direct amplification of the naive transport direction
$\mathcal{R}_{\mathrm{naive}}$, the target attractor in ACT depends on the
current editing state through the conditional denoising predictions.
Therefore, ACT produces a state-dependent transport field with adaptive
target attraction.

\paragraph{Target attraction interpretation.}

ACT can be decomposed into a naive semantic transport component and an
additional target attraction component:
\begin{align}
\hat{u}^{\mathrm{ACT}}_{k}
&=
(1+k)\hat{\mu}_{\mathrm{tar}}
-\hat{\mu}_{\mathrm{src}}
-x
\notag\\
&=
(\hat{\mu}_{\mathrm{tar}}
-\hat{\mu}_{\mathrm{src}})
+
(k\hat{\mu}_{\mathrm{tar}}-x)
\notag\\
&=
\mathcal{R}_{\mathrm{naive}}
+
a_k ,
\end{align}

where $a_k = k\hat{\mu}_{\mathrm{tar}}-x$ denotes the conditional target attraction introduced by ACT. 
This decomposition shows that ACT preserves the original semantic transition
between source and target conditions while introducing an additional
state-dependent attraction toward the target prediction. Unlike scalar
scaling of $\mathcal{R}_{\mathrm{naive}}$, the additional attraction term
depends jointly on the target conditional prediction and the current editing
state.

\begin{figure*}[h]
\centering
\includegraphics[width=\textwidth]{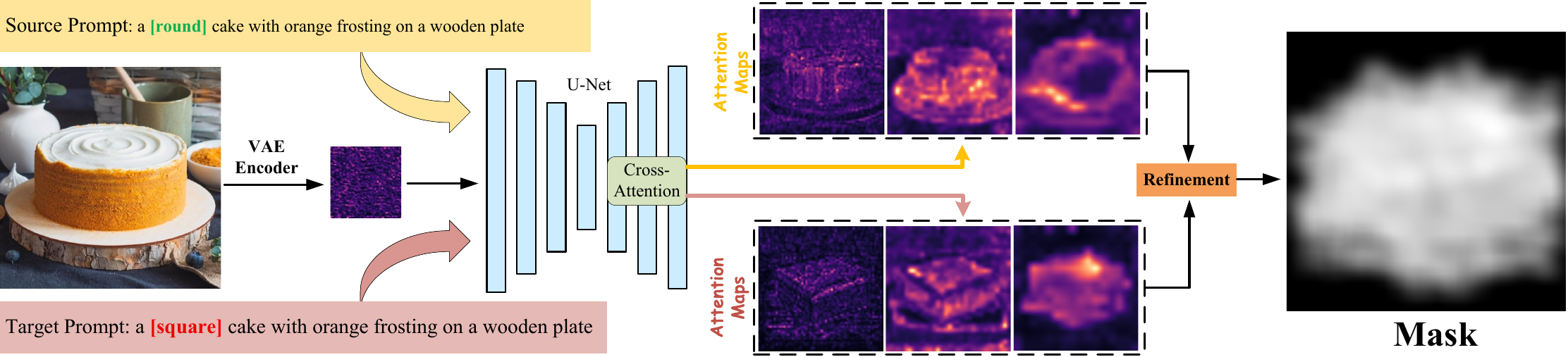}
\caption{\textbf{Attention-guided automatic local edit mask generation.}
We extract cross-attention maps from the U-Net using the changed tokens
identified from the source and target prompts. The attention responses from
inserted target tokens and removed source tokens are fused to localize the
editing region. After peak-based refinement and spatial smoothing, a soft
mask is obtained and applied to the ACT field for localized semantic
transport.}
\label{fig:Automask}
\end{figure*}

\begin{figure}[t]
\centering
\includegraphics[width=\columnwidth]{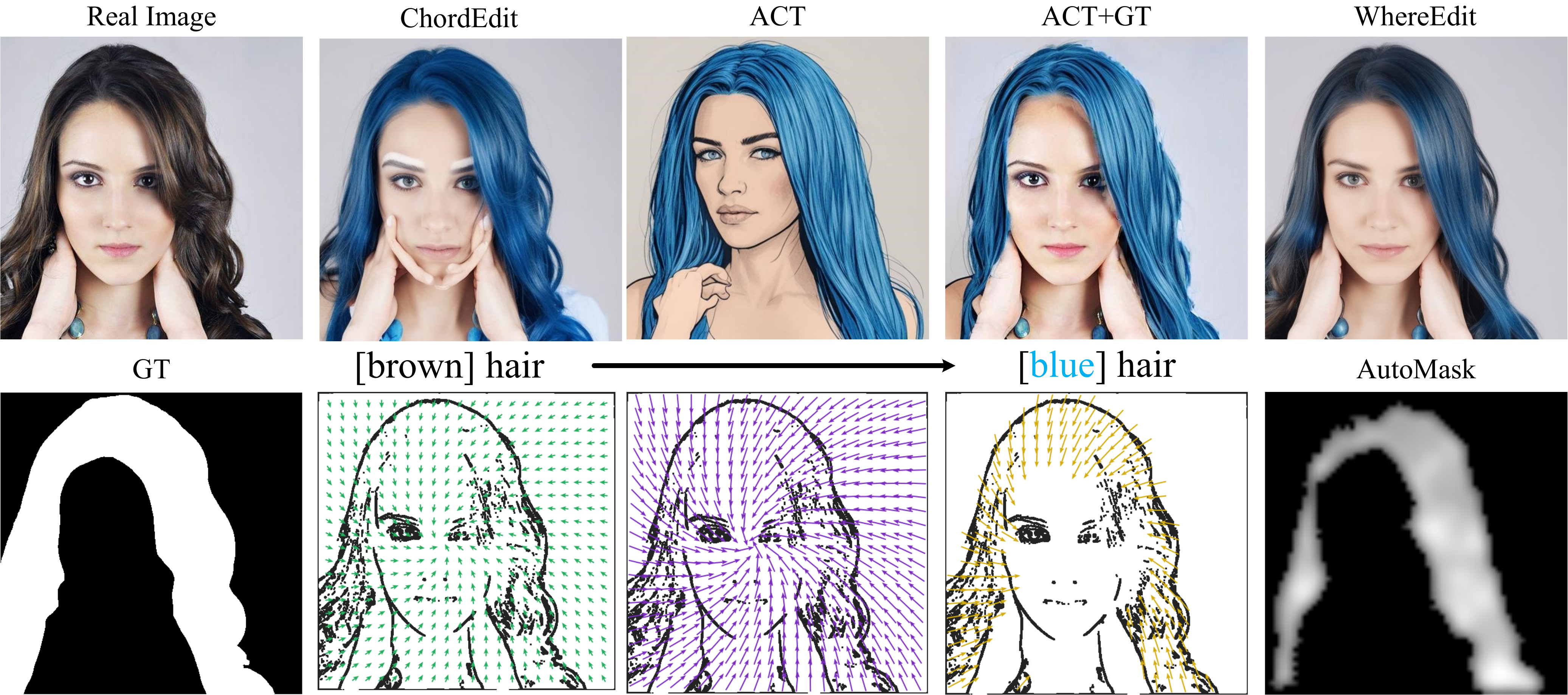}
\caption{Toy illustration of ACT and AutoMask.
ChordEdit produces a low-energy transport direction (green arrow), while ACT
strengthens target attraction with a higher-energy transport direction
(purple arrow).
AutoMask further localizes the transport field (yellow arrow), enabling
targeted edits while preserving unrelated regions.}
\label{fig:toy}
\end{figure}

To maintain temporal consistency under large amplification strengths, we
further apply a lightweight directional consistency constraint. Given two
neighboring-time ACT fields $\hat{u}_{t}$ and
$\hat{u}_{t-\delta}$ computed using the same noise bank, we define
\begin{equation}
\gamma_{t,\delta}
=
\max
\left(
0,
\cos(
\hat{u}_{t},
\hat{u}_{t-\delta}
)
\right).
\end{equation}

The final ACT update field is obtained as
\begin{equation}
\hat{u}^{\mathrm{ACT}}
=
\gamma_{t,\delta}
\hat{u}^{\mathrm{ACT}}_{k,t}.
\end{equation}

Since
$0\leq\gamma_{t,\delta}\leq1$, this constraint only suppresses temporally
inconsistent directions without increasing the update magnitude, providing
improved stability while preserving the stronger target-oriented attraction
introduced by ACT.




\subsection{Attention-Guided Automatic Local Edit Mask}
\label{sec:local-mask}

In Fig.\ref{fig:toy}, although ACT provides an effective semantic
editing direction, applying it globally may alter irrelevant regions.
We therefore introduce an automatic local edit mask module to spatially
constrain the transport field.
In Fig.~\ref{fig:Automask}, the mask is automatically generated at inference time from prompt differences and internal cross-attention, requiring no external segmentation models, annotations, or additional training while remaining lightweight.

Given source and target prompts $c_{\src}$ and $c_{\tar}$, and the latent
anchor $x$, the module predicts a soft mask
$M\in[0,1]^{1\times H\times W}$ indicating the spatial region associated
with the semantic change.
The localized ACT transport is defined as
\begin{equation}
    \hat{u}_{\mathrm{local}}
    =
    M\odot \hat{u}^{\mathrm{ACT}},
    \label{eq:local-act-mask}
\end{equation}
where $\hat{u}^{\mathrm{ACT}}$ denotes the original ACT transport field.
The latent update becomes
\begin{equation}
    x\leftarrow x+\eta\hat{u}_{\mathrm{local}}.
\end{equation}

The mask generation consists of three stages:
(1) identifying changed prompt tokens,
(2) extracting cross-attention localization maps,
and (3) refining attention responses into a compact soft mask.
An overview of how AutoMask constrains ACT transport is illustrated in
Fig.~\ref{fig:toy}.


\paragraph{Prompt-difference token localization.}

We first identify the tokens responsible for the edit.
The source and target prompts are tokenized using the frozen CLIP tokenizer
and aligned via longest-common-subsequence matching implemented by
\texttt{SequenceMatcher}.

The alignment produces edit and removal token sets:
\begin{align}
\mathcal{I}_{\mathrm{edit}}
&=
\{\text{inserted or replaced target tokens}\},\\
\mathcal{I}_{\mathrm{rem}}
&=
\{\text{deleted or replaced source tokens}\}.
\end{align}

Special tokens, including padding, BOS, EOS, and invalid IDs, are discarded.
When no explicit replacement exists, all valid target tokens are used.
This filtering focuses attention aggregation on semantic changes rather than
irrelevant prompt words.


\paragraph{Single-pass cross-attention extraction.}

We extract spatial grounding information from all UNet~\cite{ronneberger2015u, Sauer2024ADD} cross-attention layers.
Instead of additional denoising iterations, source and target conditions are
processed jointly in a single batched UNet forward pass.
At timestep $t$, the noisy latent is
\begin{equation}
z_t=\alpha_t x+\sigma_t\epsilon ,
\end{equation}
where $\epsilon$ is the shared noise used for ACT estimation.

For layer $\ell$, let
$A^{(\ell)}_{i,h}(u,v)$ denote the attention probability between spatial
location $(u,v)$ and token $i$ under head $h$.
The aggregated attention map for token set $\mathcal{I}$ is computed as
\begin{equation}
\bar A^{(\ell)}(u,v)
=
\frac{1}{|\mathcal{H}|}
\sum_{h\in\mathcal{H}}
\frac{1}{|\mathcal{I}|}
\sum_{i\in\mathcal{I}}
A^{(\ell)}_{i,h}(u,v).
\end{equation}

Attention maps from different UNet resolutions are resized to the latent
resolution and averaged across layers:
\begin{equation}
A(u,v)
=
\frac{1}{L}
\sum_{\ell=1}^{L}
\bar A^{(\ell)}(u,v).
\end{equation}

This produces two localization maps:
$A_{\mathrm{edit}}$ from target edit tokens and
$A_{\mathrm{rem}}$ from removed source tokens.



\paragraph{Source-target attention fusion.}

The new and removed concepts provide complementary localization cues.
We combine them using element-wise maximum:
\begin{equation}
A_{\mathrm{subj}}(u,v)
=
\max
(A_{\mathrm{edit}}(u,v),A_{\mathrm{rem}}(u,v)).
\end{equation}

The attention map is normalized as
\begin{equation}
\tilde A(u,v)
=
\frac{A_{\mathrm{subj}}(u,v)}
{\max_{u,v}A_{\mathrm{subj}}(u,v)+\epsilon}.
\end{equation}



\paragraph{Peak-confidence mask refinement.}

Raw cross-attention maps are spatially diffuse and may contain
background activations.
We therefore refine the normalized attention map into a compact soft mask
by extracting its peak-confidence region.
Given the normalized attention map $\tilde A$, we first locate the maximum
response:
\begin{equation}
(u^*,v^*)=\arg\max_{u,v}\tilde A(u,v).
\end{equation}

A fixed confidence threshold of $0.5$ is applied to extract the
high-response attention region:
\begin{equation}
B=
\{(u,v):
\tilde A(u,v)\geq 0.5
\}.
\end{equation}

Instead of retaining all activated regions, we keep only the connected
component containing the peak location. Let $\{C_j\}$ denote all connected
components of $B$. The selected component is defined as:
\begin{equation}
C=C_{j^*},
\quad
(u^*,v^*)\in C_{j^*}.
\end{equation}

The attention core is then obtained by masking the original attention map:
\begin{equation}
S(u,v)=C(u,v)\tilde A(u,v).
\end{equation}

To obtain smooth boundaries, we apply a small fixed morphological expansion
followed by Gaussian feathering:
\begin{equation}
M=
\frac{
\mathrm{Blur}(\mathrm{Dilate}(S))
}
{
\max_{u,v}\mathrm{Blur}(\mathrm{Dilate}(S))
}.
\end{equation}

The resulting mask preserves the high-confidence semantic region while
providing smooth transitions for latent transport.
All refinement parameters are fixed and shared across samples, avoiding
additional optimization or object-specific tuning.


\paragraph{Latent compositing.}

During editing, the ACT transport is spatially modulated by the mask:
\begin{equation}
x_{t+\Delta t}
=
x_t+\eta(M\odot\hat u^{\mathrm{ACT}}).
\end{equation}

Following ChordEdit~\cite{lu2026chordedit}, we apply target-conditioned proximal
refinement to obtain $x_{\mathrm{prox}}$.
The final result is composited as:
\begin{equation}
x_{\mathrm{out}}
=
x_{\src}\odot(1-M)
+
x_{\mathrm{prox}}\odot M.
\end{equation}

This background-preserving compositing keeps non-edited regions unchanged.
The proposed module requires only one additional UNet forward pass for
attention extraction and introduces no optimization or extra parameters.




\section{Experiment}

\paragraph{Dataset and metrics.} Following ChordEdit~\cite{lu2026chordedit}, we evaluate our method on the PIE-bench~\cite{ju2023direct} benchmark, which contains 700 instruction-based image editing samples across 10 categories at $512\times512$ resolution. Each sample includes a source image, an editing prompt, and a GT mask specifying the edited region. We evaluate performance from two perspectives: background fidelity and semantic alignment. Background fidelity is measured by PSNR and MSE on non-edited regions, while semantic alignment is evaluated using CLIP-Whole and CLIP-Edited scores~\cite{radford2021learning} for the entire image and edited region.

\paragraph{Implementation details.}
All experiments for WhereEdit are conducted using the SD-Turbo model on a
single NVIDIA RTX 3090 GPU with 24GB memory.
We adopt the single-step inference setting with the initial diffusion time
$t=1.0$.
For ACT, the target attraction strength is set to $k=2.0$, and the
neighboring-time offset for directional consistency estimation is fixed to
$\delta=0.15$.
The Monte-Carlo expectation is approximated using a single shared noise
sample ($n=1$) for all experiments.
We also evaluate the transport-only variant of ACT in
Table~\ref{tab:main_comparison_final}, where the AutoMask module is removed
to isolate the effect of spatial transport regularization.

\begin{table*}[t]
\centering
\caption{\textbf{Quantitative comparison on PIE-bench}~\cite{ju2023direct}.
\textbf{T-free}: Training-free. \textbf{I-free}: Inversion-free.
The best/second/third results in each numeric column are highlighted with
\colorbox{RankFirst}{yellow}/\colorbox{RankSecond}{orange}/\colorbox{RankThird}{blue} backgrounds, respectively.
A comprehensive table with extended metrics (e.g., SSIM, Structure Distance) is available in Appendix.}
\label{tab:main_comparison_final}
\resizebox{\textwidth}{!}{%
\begin{tabular}{@{}l l | ccc | cc | cc | ccc@{}}
\toprule
\multirow{2}{*}{\textbf{Type}} & \multirow{2}{*}{\textbf{Method}} &
\multicolumn{3}{c|}{\textbf{Consistency}} &
\multicolumn{2}{c|}{\textbf{CLIP Semantics}} &
\multicolumn{2}{c|}{\textbf{Properties}} &
\multicolumn{3}{c}{\textbf{Efficiency}} \\
\cmidrule(lr){3-5} \cmidrule(lr){6-7} \cmidrule(lr){8-9} \cmidrule(lr){10-12}
& &
\textbf{PSNR}$\uparrow$ &
\textbf{MSE}${}_{\text{10}^3}\downarrow$ &
\textbf{LPIPS}${}_{\text{10}^3}\downarrow$ &
\textbf{Whole}$\uparrow$ &
\textbf{Edited}$\uparrow$ &
\textbf{T-free} &
\textbf{I-free} &
\textbf{Step}$\downarrow$ &
\textbf{Runtime(s)}$\downarrow$ &
\textbf{VRAM(MiB)}$\downarrow$ \\
\midrule

\multirow{3}{*}{\shortstack[c]{Multi-step \\ ($\ge$ 30 steps)}} &
DDIM + MasaCtrl~\cite{song2020denoising,cao2023masactrl} &
22.19 & 8.64 & 105.44 & 24.02 & 21.14 & \cmark & \xmark &
50 & 24.49 & 13016 \\

& DDIM + PnP~\cite{song2020denoising,tumanyan2023plug} &
18.84 & 18.42 & 183.97 & 24.97 & 21.87 & \cmark & \xmark &
50 & 16.57 & 16494 \\

& FlowEdit (SD3)~\cite{kulikov2025flowedit} &
22.18 & 8.74 & 105.20 &
\cellcolor{RankSecond}26.81 &
\cellcolor{RankSecond}23.77 &
\cmark & \cmark &
33  & 6.94 & 18536 \\

\midrule

\multirow{3}{*}{\shortstack[c]{Few-step \\ (4 steps)}} &
TurboEdit (SDXL-Turbo)~\cite{deutch2024turboedit} &
22.50 & 9.30 & 105.57 & 24.68 & 21.79 &
\cmark & \cmark &
4  & 0.96 & 18004 \\

& InfEdit (SD1.4)~\cite{xu2023inversion} &
\cellcolor{RankFirst}29.59 &
\cellcolor{RankFirst}2.26 &
\cellcolor{RankFirst}37.10 &
23.04 & 20.34 &
\cmark & \cmark &
4  & 0.85 & 7620 \\

& InstantEdit (PeRFlow-SD1.5)~\cite{gong2025instantedit} &
\cellcolor{RankSecond}28.60 &
\cellcolor{RankSecond}2.80 &
\cellcolor{RankSecond}44.94 &
25.09 & 21.97 &
\cmark & \xmark &
4  & 1.45 & 10994 \\

\midrule

\multirow{4}{*}{\shortstack[c]{One-step}} &
SwiftEdit (SwiftBrush-v2)~\cite{nguyen2025swiftedit} &
24.22 & \cellcolor{RankThird}5.41 &
\cellcolor{RankThird}76.39 &
24.48 & 21.56 &
\xmark & \xmark &
\cellcolor{RankFirst}1 &
0.92 & 19628 \\

& ChordEdit (SD-Turbo) (\cite{li2026rethinking, lu2026chordedit})&
22.64 & 8.00 & 118.50 &
24.82 & 22.16 &
\cmark & \cmark &
\cellcolor{RankFirst}1 &
\cellcolor{RankSecond}0.50 &
\cellcolor{RankFirst}7418 \\

& \textbf{ACT( ours)} &
16.34 & 29.88 & 258.48 &
\cellcolor{RankFirst}27.09 &
\cellcolor{RankFirst}24.36 &
\cmark & \cmark &
\cellcolor{RankFirst}1 &
\cellcolor{RankSecond}0.50 &
\cellcolor{RankThird}7422 \\

& \textbf{WhereEdit ( ours)} &
23.96 &
8.05 &
108.91 &
\cellcolor{RankThird}25.21 &
\cellcolor{RankThird}22.66 &
\cmark & \cmark &
\cellcolor{RankFirst}1 &
0.59 &
7436 \\

\midrule

\rowcolor{gray!20}
& WhereEdit (w/ GT,  ours) &
30.21 & 1.61 & 30.92 &
25.67 & 23.63 &
\cmark & \cmark &
1  & 0.50 & 7422 \\

\bottomrule
\end{tabular}%
}
\end{table*}

\begin{figure*}[t]
\centering
\includegraphics[width=0.8\textwidth]{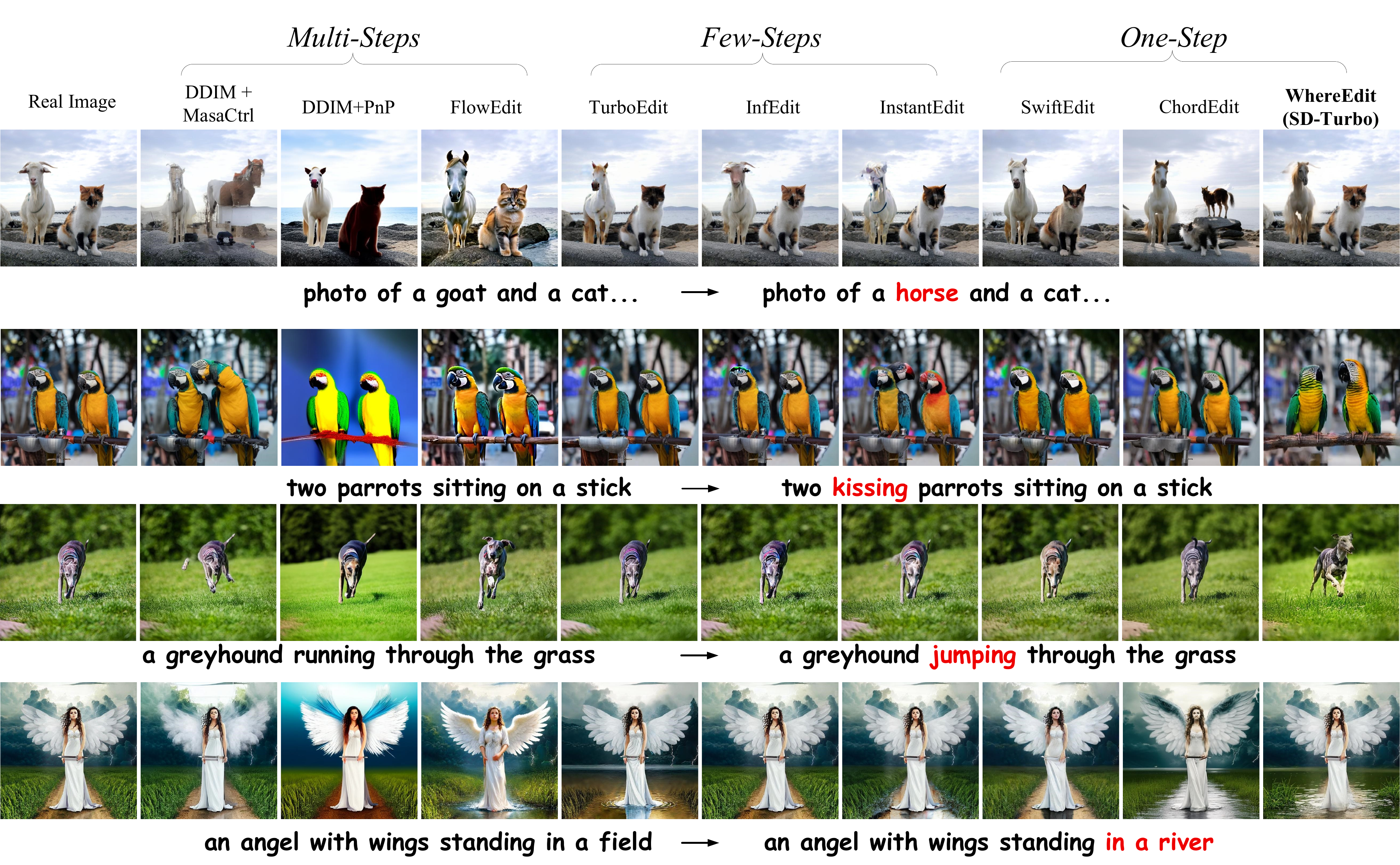}
\caption{Comparison of edited results. Real images are in the first column. Prompts are noted under each row.}
\label{fig:compare_sota}
\end{figure*}

\subsection{Comparison with Prior Methods}

As shown in Table~\ref{tab:main_comparison_final}, in the One-step image editing setting, experiments demonstrate that one-step generation can achieve competitive editing performance with extremely low inference costs. Among them, ACT achieves the best semantic editing performance, with CLIP Whole and Edited scores of 27.09 and 24.36, respectively, surpassing the multi-step FlowEdit (26.81/23.77), demonstrating that one-step frameworks can effectively balance efficient sampling and semantic controllability. Furthermore, comparing ACT with the full WhereEdit reveals that the introduced Automask module only incurs 14 MiB additional VRAM usage and 0.09s runtime overhead, indicating that it is a lightweight and efficient region localization mechanism. With Automask guidance, WhereEdit further surpasses ChordEdit, achieving a better trade-off between semantic consistency and overall editing quality. Moreover, when provided with GT, WhereEdit achieves further improvements, validating the importance of accurate region constraints for local image editing.
In Fig~\ref{fig:compare_sota}, qualitative comparisons between WhereEdit and other methods further demonstrate its superiority. Benefiting from the Automask, WhereEdit can more accurately focus on the target editing regions while preserving the structural consistency of non-edited areas, resulting in high-quality edits that better align with the given textual instructions.

\subsection{Ablation Study}
\paragraph{Analysis of ACT Coefficient $k$.}

Table~\ref{tab:act_k_ablation} studies the effect of the amplification coefficient $k$ in ACT. Increasing $k$ consistently improves the CLIP-Edited score, from 21.92 at $k=0.5$ to 23.10 at $k=4.0$, demonstrating that stronger target attraction enhances semantic alignment. However, this improvement comes with a decrease in PSNR, from 25.65 to 21.59, indicating a trade-off between editing strength and image fidelity. We set $k=2.0$ as the default, achieving a good balance with a CLIP-Edited score of 22.67 and a PSNR of 23.96. These results show that ACT benefits from an appropriate amplification strength to improve semantic editing while preserving structural consistency.

\begin{table}[t]
\centering
\scriptsize
\setlength{\tabcolsep}{3pt}

\caption{Ablation study of the ACT coefficient $k$.}
\label{tab:act_k_ablation}

\begin{tabularx}{\columnwidth}{c|*{8}{>{\centering\arraybackslash}X}}
\hline
$k$
& 0.5
& 1.0
& 1.5
& \textbf{2.0}
& 2.5
& 3.0
& 3.5
& 4.0 \\
\hline

PSNR$\uparrow$
&25.65
&25.33
&24.74
&\textbf{23.96}
&23.24
&22.62
&22.08
&21.59 \\

CLIP-Edited$\uparrow$
&21.92
&22.17
&22.43
&\textbf{22.67}
&22.81
&22.95
&23.01
&23.10 \\

\hline
\end{tabularx}

\end{table}

\begin{figure*}[h]
\centering
\includegraphics[width=\textwidth]{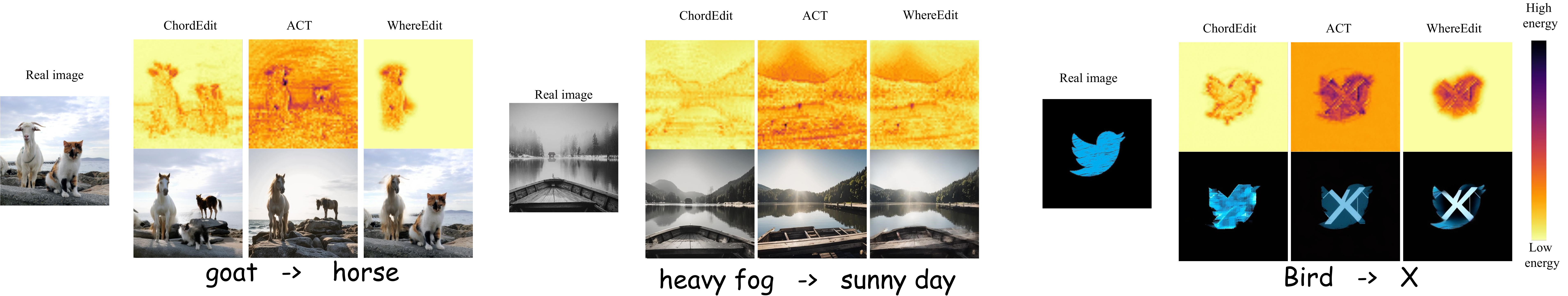}
\caption{We visualize the transport energy and editing results of ChordEdit, ACT, and
WhereEdit (ACT + AutoMask) on three examples involving local and global
semantic changes. The visualization highlights the differences in editing
strength, spatial controllability, and structural preservation among these
methods.}
\label{fig:energy_map}
\end{figure*}

\paragraph{Analysis of Editing Energy}

We analyze the editing behavior of ACT and WhereEdit from the perspective of
transport energy. We define the transport energy as
$E=\frac{1}{SC}\sum_{s=1}^{S}\sum_{\mathrm{channel}}(\hat{u}_{t}^{s})^{2}$,
where $S$ denotes the number of editing steps and $C$ is the channel
dimension. This formulation preserves the spatial structure of the transport
field for energy visualization.
As shown in Figure~\ref{fig:energy_map}, ChordEdit produces a globally
distributed low-energy transport field, which improves consistency but limits
large semantic changes. ACT introduces stronger transport energy, enabling
more expressive editing, but may affect irrelevant regions due to its global
nature. To address this issue, WhereEdit applies AutoMask to localize the
high-energy transport to semantically relevant regions, preserving ACT's
editing strength while improving spatial controllability and image fidelity.

\paragraph{Analysis of Transport and Mask Strategy.}

Table~\ref{tab:mask_comparison} investigates the interaction between transport strategies and spatial masks. Explicit masks significantly improve reconstruction fidelity by reducing unwanted modifications. However, limited CLIP-Edited gains indicate that masks alone cannot enhance semantic editing without effective transport. Compared with ChordEdit, WhereEdit achieves higher CLIP-Edited scores under all settings, demonstrating stronger semantic transport. With AutoMask, WhereEdit better localizes transport regions and balances editing strength and fidelity. Although AutoMask is slightly inferior to GT Mask, it consistently improves over the unmasked setting without requiring annotations, validating the effectiveness of adaptive localization.

\begin{table}[t]
\centering
\caption{Comparison of methods with mask strategies.}
\label{tab:mask_comparison}

\small
\renewcommand{\arraystretch}{0.75}

\resizebox{\linewidth}{!}{
\begin{tabular}{c|c|cc}
\hline
\textbf{Method} & \textbf{Mask Strategy} & \textbf{PSNR $\uparrow$} & \textbf{CLIP-Edited $\uparrow$} \\
\hline

\multirow{3}{*}{Naive} 
& w/o Mask & 20.13 & 22.57 \\
& w/ GT Mask & 31.21 & 22.23 \\
& w/ AutoMask & 26.64 & 21.29 \\

\hline

\multirow{3}{*}{ChordEdit} 
& w/o Mask & 22.64 & 22.16 \\
& w/ GT Mask & 31.38 & 21.77 \\
& w/ AutoMask & 27.89 & 21.03 \\

\hline

\multirow{3}{*}{WhereEdit} 
& w/o Mask & 16.34 & 24.36 \\
& w/ GT Mask & 30.21 & 23.63 \\
& w/ AutoMask & 23.96 & 22.66 \\

\hline
\end{tabular}
}

\end{table}

\begin{figure}[h]
\centering
\includegraphics[width=\columnwidth]{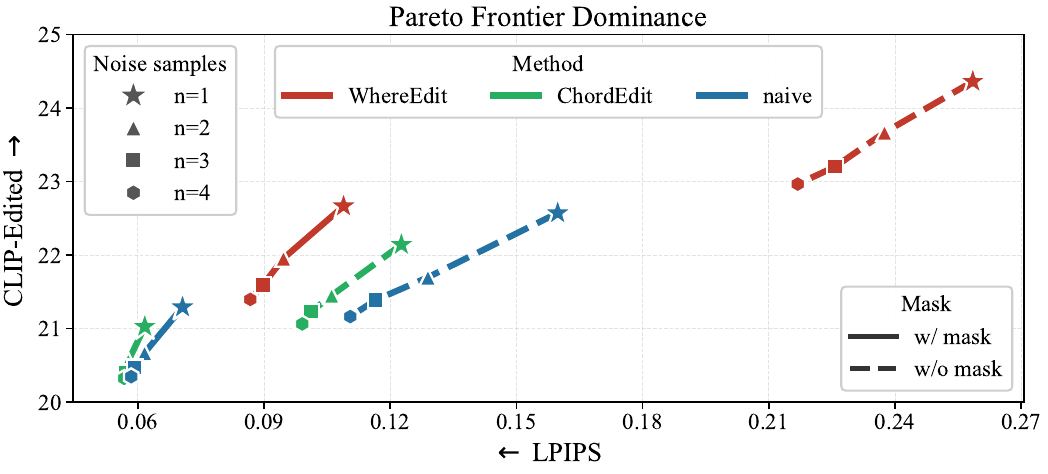}
\caption{\textbf{Effect of noise sampling and mask guidance.}
Comparison of ChordEdit, naive, and WhereEdit under different noise sample
numbers ($n$) with and without masks. More samples improve LPIPS but slightly
decrease CLIP-Edited scores. Mask guidance enhances fidelity by restricting
the editing region, while WhereEdit achieves stronger semantic editing and a
better fidelity--editing trade-off.}
\label{fig:noise}
\end{figure}

\paragraph{Analysis of Noise}

Based on Fig.~\ref{fig:noise}, increasing noise samples consistently reduces LPIPS scores, improving perceptual consistency, while slightly lowering CLIP-Edited scores due to more conservative reconstruction. With masks, WhereEdit achieves the largest improvement, reducing LPIPS from 0.259 to 0.109 at $n=1$ (57.9\% reduction), demonstrating that spatial constraints effectively reduce noise uncertainty and enhance editing consistency. Compared with sample-based stabilization, WhereEdit benefits more from region-level guidance, achieving stronger spatial controllability and robustness.

\paragraph{Analysis of Mask method}
Table~\ref{tab:mask_comparison_LIME} compares AutoMask with LIME-based localization. 
LIME achieves slightly higher PSNR, suggesting stronger background preservation, but its lower CLIP-Edited score indicates that the localized region may be overly conservative and insufficient for complete semantic modification. 
In contrast, AutoMask provides better semantic alignment while maintaining comparable fidelity and lower runtime, achieving a better trade-off between editability and preservation.

\begin{table}[t]
\centering
\caption{Comparison between ACT with Automask and LIME mask on image editing performance and efficiency. }
\label{tab:mask_comparison_LIME}
\resizebox{\linewidth}{!}{
\begin{tabular}{lccccc}
\toprule
Method & PSNR $\uparrow$ & CLIP-Edited $\uparrow$ & Runtime (s) $\downarrow$ & VRAM (MiB) $\downarrow$ & IoU (\%) $\uparrow$ \\
\midrule
Automask & 23.96 & 22.66 & 0.59 & 7436 & 44.19 \\
LIME~\cite{Simsar_2025_WACV}     & 24.65 & 21.21 & 0.72 & 7450 & 16.59\\
\bottomrule
\end{tabular}
}
\end{table}

\section{Conclusion}
In this work, we investigate the fundamental challenges of one-step image
editing: how to automatically identify editable regions and how to achieve
strong semantic modification under a single-step generation paradigm.
We propose WhereEdit, a training-free and inversion-free framework that
reformulates one-step editing as a spatially controlled semantic transport
problem. By introducing an attention-guided AutoMask mechanism, WhereEdit
automatically discovers editing regions from token-level attention responses,
while Amplified Conditional Transport provides stronger and more stable
semantic transformation within the localized area.
Extensive experiments on PIE-Bench demonstrate that WhereEdit achieves
superior editing quality among one-step editing methods while maintaining the
efficiency advantages of fast generation. The improvements obtained with
ground-truth masks further highlight the importance of explicit spatial
reasoning for high-quality one-step editing.
We believe WhereEdit provides a
practical direction toward real-time, controllable, and high-fidelity image
editing, bridging the gap between efficient one-step generation and precise
spatial manipulation.

\bibliography{aaai2027}

@inproceedings{song2023consistency,
  title={Consistency Models},
  author={Song, Yang and Dhariwal, Prafulla and Chen, Mark and Sutskever, Ilya},
  booktitle={International Conference on Machine Learning},
  pages={32211--32252},
  year={2023},
  organization={PMLR}
}

@inproceedings{liu2023instaflow,
  title={Instaflow: One step is enough for high-quality diffusion-based text-to-image generation},
  author={Liu, Xingchao and Zhang, Xiwen and Ma, Jianzhu and Peng, Jian and Liu, Qiang},
  booktitle={International Conference on Learning Representations},
  year={2024}
}

@article{hertz2022prompt,
  title = {Prompt-to-Prompt Image Editing with Cross Attention Control},
  author = {Hertz, Amir and Mokady, Ron and Tenenbaum, Jay and Aberman, Kfir and Pritch, Yael and Cohen-Or, Daniel},
  journal = {arXiv preprint arXiv:2208.01626},
  year = {2022},
}

@inproceedings{cao2023masactrl,
  title={Masactrl: Tuning-free mutual self-attention control for consistent image synthesis and editing},
  author={Cao, Mingdeng and Wang, Xintao and Qi, Zhongang and Shan, Ying and Qie, Xiaohu and Zheng, Yinqiang},
  booktitle={Proceedings of the IEEE/CVF international conference on computer vision},
  pages={22560--22570},
  year={2023}
}

@inproceedings{tumanyan2023plug,
  title={Plug-and-play diffusion features for text-driven image-to-image translation},
  author={Tumanyan, Narek and Geyer, Michal and Bagon, Shai and Dekel, Tali},
  booktitle={Proceedings of the IEEE/CVF conference on computer vision and pattern recognition},
  pages={1921--1930},
  year={2023}
}

@inproceedings{mokady2023null,
  title={Null-text inversion for editing real images using guided diffusion models},
  author={Mokady, Ron and Hertz, Amir and Aberman, Kfir and Pritch, Yael and Cohen-Or, Daniel},
  booktitle={Proceedings of the IEEE/CVF conference on computer vision and pattern recognition},
  pages={6038--6047},
  year={2023}
}

@inproceedings{kulikov2025flowedit,
  title={Flowedit: Inversion-free text-based editing using pre-trained flow models},
  author={Kulikov, Vladimir and Kleiner, Matan and Huberman-Spiegelglas, Inbar and Michaeli, Tomer},
  booktitle={Proceedings of the IEEE/CVF International Conference on Computer Vision},
  pages={19721--19730},
  year={2025}
}

@inproceedings{nguyen2025swiftedit,
  title={Swiftedit: Lightning fast text-guided image editing via one-step diffusion},
  author={Nguyen, Trong-Tung and Nguyen, Quang and Nguyen, Khoi and Tran, Anh and Pham, Cuong},
  booktitle={Proceedings of the Computer Vision and Pattern Recognition Conference},
  pages={21492--21501},
  year={2025}
}

@inproceedings{deutch2024turboedit,
  title={Turboedit: Text-based image editing using few-step diffusion models},
  author={Deutch, Gilad and Gal, Rinon and Garibi, Daniel and Patashnik, Or and Cohen-Or, Daniel},
  booktitle={SIGGRAPH Asia 2024 Conference Papers},
  pages={1--12},
  year={2024}
}

@article{xu2023inversion,
  title={Inversion-Free Image Editing with Natural Language},
  author={Xu, Sihan and Huang, Yidong and Pan, Jiayi and Ma, Ziqiao and Chai, Joyce},
  journal={CoRR},
  year={2023}
}

@inproceedings{gong2025instantedit,
  title={InstantEdit: Text-Guided Few-Step Image Editing with Piecewise Rectified Flow},
  author={Gong, Yiming and Zhu, Zhen and Zhang, Minjia},
  booktitle={Proceedings of the IEEE/CVF International Conference on Computer Vision},
  pages={16808--16817},
  year={2025}
}

@article{ju2023direct,
  title={PnP Inversion: Boosting Diffusion-based Editing with 3 Lines of Code},
  author={Ju, Xuan and Zeng, Ailing and Bian, Yuxuan and Liu, Shaoteng and Xu, Qiang},
  journal={International Conference on Learning Representations ({ICLR})},
  year={2024}
}

@article{icd24,
  author  = {Nikita Starodubcev and Mikhail Khoroshikh and Artem Babenko and Dmitry Baranchuk},
  title   = {Invertible Consistency Distillation for Text-Guided Image Editing in Around 7 Steps},
  journal = {arXiv:2406.14539},
  year    = {2024},
  url     = {https://arxiv.org/abs/2406.14539}
}

@inproceedings{edict23,
  author    = {Bram Wallace and Akash Gokul and Nikhil Naik},
  title     = {EDICT: Exact Diffusion Inversion via Coupled Transformations},
  booktitle = {CVPR},
  year      = {2023},
  url       = {https://openaccess.thecvf.com/content/CVPR2023/papers/Wallace_EDICT_Exact_Diffusion_Inversion_via_Coupled_Transformations_CVPR_2023_paper.pdf}
}

@inproceedings{aidi23,
  author    = {Zhizhong Pan and Yijun Li and Xue Bai and Zhuowen Tu and Ming-Hsuan Yang},
  title     = {Effective Real Image Editing with Accelerated Iterative Diffusion Inversion},
  booktitle = {ICCV},
  year      = {2023},
  url       = {https://openaccess.thecvf.com/content/ICCV2023/html/Pan_Effective_Real_Image_Editing_with_Accelerated_Iterative_Diffusion_Inversion_ICCV_2023_paper.html}
}

@inproceedings{npi25,
  author    = {Daiki Miyake and Akihiro Iohara and Yu Saito and Toshiyuki Tanaka},
  title     = {Negative-Prompt Inversion: Fast Image Inversion for Editing with Text-Guided Diffusion Models},
  booktitle = {WACV},
  year      = {2025},
  url       = {https://openaccess.thecvf.com/content/WACV2025/html/Miyake_Negative-Prompt_Inversion_Fast_Image_Inversion_for_Editing_with_Text-Guided_Diffusion_WACV_2025_paper.html}
}

@inproceedings{proxedit24,
  author    = {Bo Han and Yujun Shen and Yingqing He and Lei Zhang and Jingren Zhou},
  title     = {ProxEdit: Improving Tuning-Free Real Image Editing with Proximal Guidance},
  booktitle = {WACV},
  year      = {2024},
  url       = {https://openaccess.thecvf.com/content/WACV2024/papers/Han_ProxEdit_Improving_Tuning-Free_Real_Image_Editing_With_Proximal_Guidance_WACV_2024_paper.pdf}
}

@inproceedings{spdinv24,
  author    = {Ruibin Li and Yujun Shen and others},
  title     = {Source Prompt Disentangled Inversion for Boosting Image Editability with Diffusion Models},
  booktitle = {ECCV},
  year      = {2024},
  url       = {https://www.ecva.net/papers/eccv_2024/papers_ECCV/papers/03804.pdf}
}

@inproceedings{Sauer2024ADD,
  author    = {Axel Sauer and Dominik Lorenz and Andreas Blattmann and Robin Rombach},
  title     = {Adversarial Diffusion Distillation},
  booktitle = {Computer Vision -- ECCV 2024},
  series    = {Lecture Notes in Computer Science},
  volume    = {15144},
  pages     = {87--103},
  publisher = {Springer},
  year      = {2024},
  doi       = {10.1007/978-3-031-73016-0_6},
  url       = {https://arxiv.org/abs/2311.17042}
}

@incollection{Dao2025SwiftBrushV2,
  author    = {Trung Dao and Thuan Hoang Nguyen and Thanh Le and Duc Vu and Khoi Nguyen and Cuong Pham and Anh Tran},
  title     = {SwiftBrush V2: Make Your One-Step Diffusion Model Better Than Its Teacher},
  booktitle = {Computer Vision -- ECCV 2024},
  series    = {Lecture Notes in Computer Science},
  volume    = {15140},
  pages     = {176--192},
  publisher = {Springer},
  year      = {2025},   
  doi       = {10.1007/978-3-031-73007-8_11},
  url       = {https://arxiv.org/abs/2408.14176}
}

@inproceedings{kawar2023imagic,
  title     = {Imagic: Text-Based Real Image Editing with Diffusion Models},
  author    = {Kawar, Bahjat and Zada, Shiran and Lang, Oran and Tov, Omer and Chang, Huiwen and Dekel, Tali and Mosseri, Inbar and Irani, Michal},
  booktitle = {Proceedings of the IEEE/CVF Conference on Computer Vision and Pattern Recognition (CVPR)},
  year      = {2023},
  url       = {https://openaccess.thecvf.com/content/CVPR2023/papers/Kawar_Imagic_Text-Based_Real_Image_Editing_With_Diffusion_Models_CVPR_2023_paper.pdf}
}

@inproceedings{radford2021learning,
  title={Learning transferable visual models from natural language supervision},
  author={Radford, Alec and Kim, Jong Wook and Hallacy, Chris and Ramesh, Aditya and Goh, Gabriel and Agarwal, Sandhini and Sastry, Girish and Askell, Amanda and Mishkin, Pamela and Clark, Jack and others},
  booktitle={International conference on machine learning},
  pages={8748--8763},
  year={2021},
  organization={PmLR}
}

@article{lin2024sdxl,
  title={Sdxl-lightning: Progressive adversarial diffusion distillation},
  author={Lin, Shanchuan and Wang, Anran and Yang, Xiao},
  journal={arXiv preprint arXiv:2402.13929},
  year={2024}
}

@inproceedings{lu2026chordedit,
  title={Chordedit: One-step low-energy transport for image editing},
  author={Lu, Liangsi and Chen, Xuhang and Guo, Minzhe and Li, Shichu and Wang, Jingchao and Shi, Yang},
  booktitle={Proceedings of the IEEE/CVF Conference on Computer Vision and Pattern Recognition},
  pages={14398--14407},
  year={2026}
}

@article{li2026rethinking,
  title={Rethinking One-Step Image Editing through ChordEdit: Reproduction, Simplification, and New Insights},
  author={Li, Minghan and Moebel, Jeremy and Wang, Mengyu},
  journal={arXiv preprint arXiv:2606.14042},
  year={2026}
}

@inproceedings{hu2026fb,
  title={FB-CLIP: Fine-Grained Zero-Shot Anomaly Detection with Foreground-Background Disentanglement},
  author={Hu, Ming and Huo, Yongsheng and Dou, Mingyu and Yin, Jianfu and Zhao, Peng and Wang, Yao and Hu, Cong and Hu, Bingliang and Wang, Quan},
  booktitle={Proceedings of the IEEE/CVF Conference on Computer Vision and Pattern Recognition},
  pages={35659--35669},
  year={2026}
}

@inproceedings{hu2025beta,
  title={beta-fft: Nonlinear interpolation and differentiated training strategies for semi-supervised medical image segmentation},
  author={Hu, Ming and Yin, Jianfu and Ma, Zhuangzhuang and Ma, Jianheng and Zhu, Feiyu and Wu, Bingbing and Wen, Ya and Wu, Meng and Hu, Cong and Hu, Bingliang and others},
  booktitle={Proceedings of the IEEE/CVF Conference on Computer Vision and Pattern Recognition},
  pages={30839--30849},
  year={2025}
}

@inproceedings{hu2024specslice,
  title={Specslice-convlstm: Medical hyperspectral image segmentation using spectral slicing and convlstm},
  author={Hu, Ming and Yin, Jianfu and Wang, Jing and Wang, Yuqi and Hu, Bingliang and Wang, Quan},
  booktitle={International Conference on Pattern Recognition},
  pages={211--225},
  year={2024},
  organization={Springer}
}

@article{ho2020denoising,
  title={Denoising diffusion probabilistic models},
  author={Ho, Jonathan and Jain, Ajay and Abbeel, Pieter},
  journal={Advances in neural information processing systems},
  volume={33},
  pages={6840--6851},
  year={2020}
}

@article{song2020denoising,
  title={Denoising diffusion implicit models},
  author={Song, Jiaming and Meng, Chenlin and Ermon, Stefano},
  journal={arXiv preprint arXiv:2010.02502},
  year={2020}
}

@article{peng2025facm,
  title={FACM: Flow-Anchored Consistency Models},
  author={Peng, Yansong and Zhu, Kai and Liu, Yu and Wu, Pingyu and Li, Hebei and Sun, Xiaoyan and Wu, Feng},
  journal={arXiv preprint arXiv:2507.03738},
  year={2025}
}

@article{geng2026mean,
  title={Mean flows for one-step generative modeling},
  author={Geng, Zhengyang and Deng, Mingyang and Bai, Xingjian and Kolter, Zico and He, Kaiming},
  journal={Advances in Neural Information Processing Systems},
  volume={38},
  pages={75460--75482},
  year={2026}
}

@InProceedings{Simsar_2025_WACV,
    author    = {Simsar, Enis and Tonioni, Alessio and Xian, Yongqin and Hofmann, Thomas and Tombari, Federico},
    title     = {LIME: Localized Image Editing via Attention Regularization in Diffusion Models},
    booktitle = {Proceedings of the Winter Conference on Applications of Computer Vision (WACV)},
    month     = {February},
    year      = {2025},
    pages     = {222-231}
}

@article{tang2024locinv,
  title={Locinv: localization-aware inversion for text-guided image editing},
  author={Tang, Chuanming and Wang, Kai and Yang, Fei and van de Weijer, Joost},
  journal={arXiv preprint arXiv:2405.01496},
  year={2024}
}

@inproceedings{mao2024mag,
  title={Mag-edit: Localized image editing in complex scenarios via mask-based attention-adjusted guidance},
  author={Mao, Qi and Chen, Lan and Gu, Yuchao and Fang, Zhen and Shou, Mike Zheng},
  booktitle={Proceedings of the 32nd ACM International Conference on Multimedia},
  pages={6842--6850},
  year={2024}
}

@inproceedings{fu2026layeredit,
  title={Layeredit: Disentangled multi-object editing via conflict-aware multi-layer learning},
  author={Fu, Fengyi and Huang, Mengqi and Zhang, Lei and Mao, Zhendong},
  booktitle={Proceedings of the AAAI Conference on Artificial Intelligence},
  volume={40},
  number={5},
  pages={4003--4011},
  year={2026}
}

@inproceedings{ronneberger2015u,
  title={U-net: Convolutional networks for biomedical image segmentation},
  author={Ronneberger, Olaf and Fischer, Philipp and Brox, Thomas},
  booktitle={International Conference on Medical image computing and computer-assisted intervention},
  pages={234--241},
  year={2015},
  organization={Springer}
}

\end{document}